\documentclass[10pt]{article}
\pdfoutput=1
\def\doctitle{From RESTful Services to RDF: Connecting the Web and the Semantic Web}
\def\docauthor{Rosa Alarcon and Erik Wilde}
\def\repdate{June 2010}
\def\irep{UC Berkeley School of Information Report 2010-041}

\usepackage{verbatim}

\usepackage{graphicx}
\usepackage{fancyheadings}
\usepackage{lastpage}
\usepackage{boxedminipage}
\usepackage[margin=1in,letterpaper,includeheadfoot]{geometry}
\usepackage[a4paper={false},letterpaper={true},
colorlinks={true},
linkcolor={blue},
citecolor={black},
urlcolor={blue},
pdfstartview={Fit},
pdfview={Fit},
bookmarks={true},
pdftitle={\doctitle},
pdfsubject={\irep},
pdfauthor={\docauthor}]{hyperref}

\def\uri#1{{\tt \url{#1}}}
\def\code#1{{\tt #1}}

\begin{document}

\title{\vspace{-1cm}From RESTful Services to RDF:\\
Connecting the Web and the Semantic Web}
\author{Rosa Alarcon $^1$ and Erik Wilde $^2$\\
$^1$ \href{http://www.dcc.uchile.cl/}{Departamento de Ciencia de la Computacion, Pontificia Universidad Catolica de Chile}\\
$^2$ \href{http://www.ischool.berkeley.edu/}{School of Information, UC Berkeley}}
\date{\irep\\
\repdate\\
\ \\
Available at \uri{http://escholarship.org/uc/item/3425p9s7}}
\maketitle

\thispagestyle{empty}

\begin{abstract}

RESTful services on the Web expose information through retrievable resource representations that represent self-describing descriptions of resources, and through the way how these resources are interlinked through the hyperlinks that can be found in those representations. This basic design of RESTful services means that for extracting the most useful information from a service, it is necessary to understand a service's representations, which means both the semantics in terms of describing a resource, and also its semantics in terms of describing its linkage with other resources. Based on the \emph{Resource Linking Language (ReLL)}, this paper describes a framework for how RESTful services can be described, and how these descriptions can then be used to harvest information from these services. Building on this framework, a layered model of RESTful service semantics allows to represent a service's information in RDF/OWL. Because REST is based on the linkage between resources, the same model can be used for aggregating and interlinking multiple services for extracting RDF data from sets of RESTful services.

\end{abstract}

\vfill
\tableofcontents
\newpage

\section{Introduction}

The core model of the \emph{Semantic Web}~\cite{ber01} is centered around resources that are identified by URIs, and by descriptions that make assertions about these resources based on properties (which also are identified by URIs) and values assigned to these properties (which can be URIs or literal values). This interconnected network of URI-described resources is defined by the \emph{Resource Description Framework (RDF)}~\cite{rdfconcepts}, and more sophisticated languages then build additional layers of semantics on this foundation. The underlying assumption of this approach is that the Web exposes a large number of resources, and that RDF can therefore be used to describe this large set of resources. In this paper, we describe how resources can be discovered using the Web's basic architectural principle (REST), and how this discovery process can be used to expose resources and their relationships as RDF data.

Large amounts of RDF interlinked data are required in order to provide a critical mass of information for developers, and reaching this critical mass is one of the initial problems of the Semantic Web vision. Therefore, there is a lot of activity in research projects that create large collections of RDF data by transforming structured data sources into RDF using specialized mappings, exposing the generated RDF dataset in RDF triple stores, often exposing query interfaces in a RDF-oriented query language such as \emph{SPARQL}~\cite{sparqlquery}. Resolvable semantic resource URIs are provided and triples provide connectivity in this graph of RDF data for steering semantic crawlers and Semantic Web browsers. Although this approach creates large collections of RDF data, they result in centralistic approaches where access is typically mediated through a single ``endpoint'' and require sophisticated mechanisms to retrieve, process, and publish the information~\cite{boj08}.


On the other hand, there is an increasing interest in the relationship of \emph{Representational State Transfer (REST)}~\cite{fie02}, the architectural principle underlying the Web, and the Semantic Web. Approaches in this area vary from the semantic annotation of resources (e.g., hREST, SA-REST), to middleware that mediates resources handling, following in general the same approaches of more traditional SOAP/WSDL semantic services (e.g., WSMO). But REST requires a different approach because services are based on the principles of resources identified with unique and opaque URIs, that are resolved to clients in various ``representations'' with a media type, and are handled through a uniform interface. REST resources are interlinked through hyperlinks that are found in these representations and guide clients in their interactions with a service.

This paper presents a metamodel for describing RESTful services and a language for creating descriptions of these services that is based on REST's central principle, the hyperlinking of resources. This approach provides a natural mapping from the graph-oriented world of RESTful services (resources interlinked by links found in resource representations) to the graph-based model of RDF. It is possible to directly expose structured data in Web services that expose an interface in line with \emph{Representational State Transfer (REST)}~\cite{fie02}, thereby supporting lightweight approaches for structured data~\cite{wil08i}. Based on this starting point of using plain Web technologies, it is possible to go one step further and expose the same data based on Semantic Web technologies. Individual resource representations as well as the hyperlinks connecting them can be mapped to RDF in a variety of ways. \emph{Gleaning Resource Descriptions from Dialects of Languages (GRDDL)}~\cite{grddl} is a framework for doing this, but as long as there are well-defined mappings, any RESTful service can be transformed into an RDF graph. This even includes representations such as images or PDF documents, which might contain embedded metadata, which can be extracted with an appropriate toolset.


\section{Related Work}\label{related}

Related work in Web services and Semantic Web can be broadly categorized into two areas. \emph{Semantic Web Services} (Section~\ref{sws}) deal with how to either annotate service descriptions with semantic annotations, or how to design and describe Web services that directly provide support for Semantic Web technologies. \emph{Harvesting} (Section~\ref{harvestingRW}) deals with the question of how to use existing Web services (often those which do not expose any Semantic Web data) in a way so that they can serve as providers for Semantic Web data.

\subsection{Semantic Web Services}\label{sws}

\emph{Semantic Web Services (SWS)} address mainly SOAP/WSDL services which, without additional annotations, are only focused on the syntax required for describing exposed functionality (operations, input and output types). SWS approaches extend WSDL services with semantic models. For instance, OWL-S~\cite{owl-s} proposes a meta-ontology describing services in terms of operations, inputs, outputs, preconditions, and effects; WSMO~\cite{wsmo} follows a similar approach though domain ontologies and goals are allowed for service discovery and composition, requiring a highly specialized reasoning platform. A lightweight approach is SAWSDL~\cite{sawsdl}, which allows annotations in WSDL descriptions referring to elements in a semantic model.

A similar approach has been proposed for RESTful services. Since REST services lack a service description, SA-REST~\cite{lat07} and hREST/MicroWSMO~\cite{kop08} propose a service description as an annotated resource (e.g., an HTML page) containing the list of input and output parameters, methods, and URIs exposed by a service by means of property value pairs or RDFa~\cite{rdfasyntax} annotations. The description is transformed to RDF using a GRDDL-based~\cite{grddl} strategy for generating a domain ontology in RDF, but no information about the REST resources themselves (instances) are retrieved. Battle and Benson~\cite{bat08} provide similar annotations to WADL~\cite{wadl} documents describing REST services, and also propose extensions to SPARQL in order to support an HTTP REST uniform interface. Extensions to the payload of the HTTP REST methods (e.g., \code{PUT}, \code{DELETE} and \code{GET}) are also proposed for keeping consistency between a REST resource and its semantic equivalence (a triple) in some triple store.

The main problem of these approaches is that they are based on the assumption of a ``service endpoint'' (which is then described semantically), so they basically reflect the RPC-style service model of WSDL/SOAP. By using a RPC-style for the description of the service, though, they do not align well with the principles of RESTful service design, since they disregard fundamental properties such as the hypermedia nature of REST, and the possibility of multiple representations for resources. They also introduce coupling in their design by adhering to URI templates for describing the URIs of resources, input, and output parameters~\cite{pau09b}, or in the case of Battle and Benson, they introduce new semantics to the standard REST interface.

EXPRESS~\cite{alo09} is a SWS model that explicitly avoids the RPC-orientation of the approaches mentioned so far. It starts from HTTP's uniform interface, and then describes the available resources in an OWL ontology. However, the model of EXPRESS is a centralized one as well, because it is assumed that there is a complete description of a Web Service's available resources, and then this description is used to generate URIs for classes, instances, and properties.

\subsection{Harvesting RDF data from Web resources}\label{harvestingRW}

As a rough classification of how to extract RDF from existing Web sources, there are approaches built specifically around using one particular dataset/service, such as DBpedia~\cite{aue07} and the growing set of other datasets exposed as linked data, and there are generic approaches. The generic approaches can be further categorized into those that extract information based on the explicit structures found in data sources; and those that utilize additional rules for information extraction, often based on \emph{Natural Language Processing (NLP)} or other information retrieval methods.


In the approach described by Futrelle~\cite{fut06}, RDF is used as the ``integration layer'' in a scenario of heterogeneous data sources, and the main focus is on harvesting well-known and cooperating data sources. This approach thus falls in the category where it can be applied to a variety of data sources, but they have to be cooperating in the sense that they expose RDF themselves. The harvester's main role is to be notified of new and updated data, and to pull it in from these sources. While this scenario uses RDF's power to unify heterogeneous data sources on the metamodel level, it is only applicable in closed and cooperating settings. In our approach, data sources are not required to publish RDF themselves. As long as access to data is provided through RESTful services, they can be harvested and used as RDF. A weakness of the current implementation is that updating is not supported in a way that allows efficient incremental updates, but we plan to address this issue in our future work mentioned in Section~\ref{conclusions}, where we describe extensions to our language that represent update services (and thus the ability to use those for incremental updates) on the language level.

SOFIE~\cite{suc09} focuses on information extraction from Web resources, and ANGIE~\cite{pre09} on using both extracted information and Web services endpoints, for building a more interactive system that does not require an exhaustive crawl of data, but retrieves information on demand. SOFIE thus falls into the category of approaches that start from resource representations, and use information retrieval methods to extract RDF from them. The current implementation of ANGIE focus on the dynamics of query processing in the RDF data managed by the system, and uses a hardwired set of Web services as the back-end. Similar to SA-REST, it uses a set of lowering/lifting transformations to translate the results of function calls from and to RDF. ANGIE focuses on SPARQL processing (the framework is able to use Web services while processing SPARQL queries), and less on the ability to easily accommodate a large variety of RESTful services.

{\sc Deimos}~\cite{amb09} is another system that starts with information found on Web pages or through Web forms, and then uses semantic analysis to map the syntax of these representations to semantically richer information. Instead of relying on the richness of links discovered in known resources, though, the approach taken in {\sc Deimos} uses tagging services to discover new resources.


\section{REST Semantics}\label{rest-semantics}

Unlike WSDL services that expose URIs identifying endpoints where service functionality can be invoked (based on the underlying RPC model), REST services expose URIs for a set of resources and clients encounter URIs by following hyperlinks. In order to avoid coupling between clients and servers, resource URIs must be opaque, that is, no assumptions must be made about the structure of the URI (a popular approach that violates this principle is that of URI templates, where URIs are composed based on a template and instantiation rules). Instead, resource URIs must be discovered by following the hyperlinks embedded in a resource representation, which ensures that clients are not tightly coupled to any particular URI structure~\cite{pau09a}.

REST requires a uniform interface that depends on the scheme used for a URI, in case of HTTP, the standard methods are \code{GET}, \code{PUT}, \code{POST}, \code{DELETE}, and \code{OPTIONS}. Methods are external to the resources, and are invoked by sending standard messages to the server indicating the URI of the requested resource, the method, the payload of the message and possibly standard metadata (HTTP header fields). These invocations may force changes in the state of the resources according to the semantics of the method. REST resources are conceptual entities that belong to the application logic, and are rendered to the user as representations that convey a standardized format or media type (e.g., \code{text/html},\code{application/xml}, etc.). The content of representations depends on the application scenario and requested resource, but in RESTful designs they must contain the necessary links that allow clients to discover other resources (or affect the state of a resource) by following such links.

These properties imply that there are no ``endpoints'' for the case of REST services, instead there is a collection of resource URIs and a set of standard operations. A resource can have multiple representations (e.g., plain text, HTML, or PDF) that can be negotiated. Resource URIs are discovered by following the links contained in representations, that is, representations contain links to related representations. The sources for semantic information differ greatly from the WSDL model: In REST graphs, resources, links and representations are fundamental, whereas in WSDL, the complexity of an exposed service is often mostly exposed as a set of available functions that must be known in advance, and then can be invoked by the client.

\section{REST Semantics in ReLL}\label{rell}

\begin{figure}[tb]
\begin{center}
\includegraphics{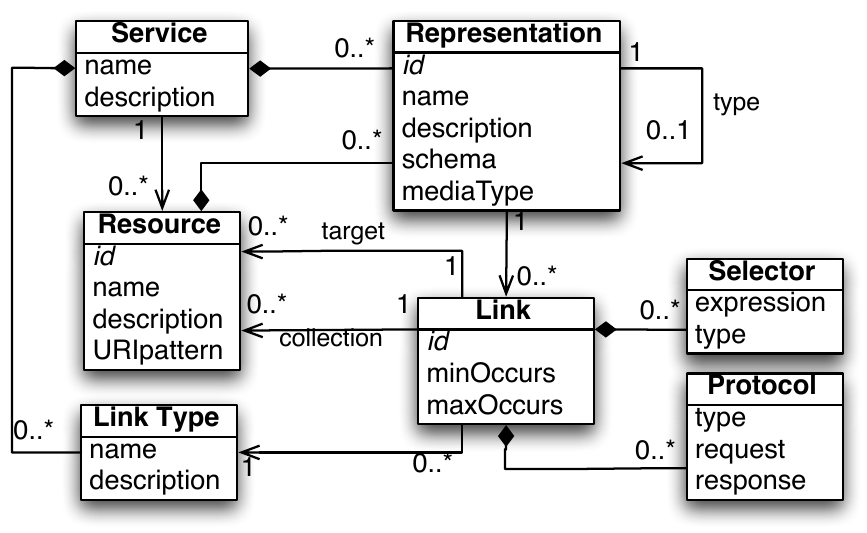}
\caption{ReLL Metamodel for the description of REST services}\label{fig-rell-model}
\end{center}
\end{figure}

Figure~\ref{fig-rell-model} shows a metamodel for REST service descriptions. This metamodel is the basis for the \emph{Resource Linking Language (ReLL)} which is a language describing interlinked REST resources, and thus the service that can be accessed by interacting with those resources. A \emph{service} provides one or more \emph{resources}, with human readable names, descriptions and a \emph{URI}. URIs are opaque and are only described as patterns (e.g., regular expressions), but even those are optional. Resources may have \emph{representations}, that is, the serialization of the resource in some syntax (e.g., HTML, Atom, etc.) and can be associated with \emph{schemas} for possible validation (of retrieved resources).

Representations may contain \emph{links}, relating the represented resource to a \emph{target} resource. A link may have a \emph{link type} defining the semantics of the link, a name, and a description. Links can be extracted from resources by using a \emph{selector}, which in case of XML-based representations is an \emph{XML Path Language (XPath)} expression that allows structured selections within XML document trees. Links can also provide additional information on how to use a specific \emph{protocol} when following the link. ReLL does not restrict the use of protocols to \emph{Hypertext Transfer Protocol (HTTP)}~\cite{rfc2616}, but for HTTP, it supports additional information about the method to be used in the request, and optionally the request payload if additional information such as query parameters or a request entity is required. Methods depend on the protocol, and thus additional methods provided by HTTP extensions such as \emph{WebDAV}~\cite{rfc4918}, \emph{CalDAV}~\cite{rfc4791} or \code{PATCH}~\cite{dus09} can also be used.

Links that do specify a URI pattern might use URIs that do not match the pattern, indicating that it leads to a resource that is outside the scope of the service description. This means, that an application such as the crawler described in Section~\ref{implementation} shall not dereference that link. This highlights the fact that it is possible to have more than one description of a service, depending on the specific interests when using that service. Thus, a ReLL description often serves a specific purpose when using a service, and thus different users of a service might want to use different ReLL descriptions of that service, excluding or constraining the service with regard to representations or links they are not interested in.

Well-known linking patterns such as collections (e.g., ``paged'' representations) are modeled as \emph{collection} links. These links represent sets of representations that correspond to a collection of resources. Collection links are not allowed to specify a \emph{target}, since the target of a collection link is the same resource that contains the collection link.

\section{Describing Services with ReLL}\label{description}

\begin{figure}[tb]
\begin{center}
\begin{boxedminipage}{\textwidth}
\includegraphics[width=\columnwidth]{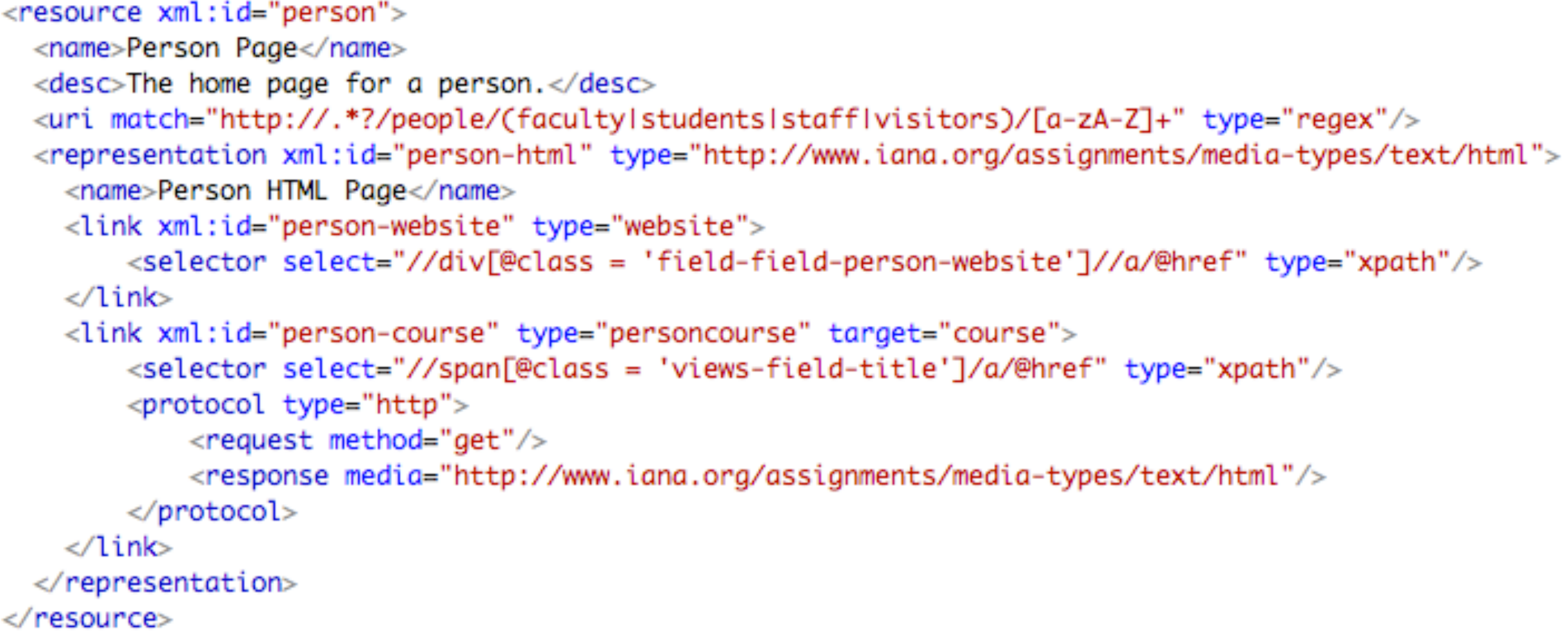}
\end{boxedminipage}
\caption{A snippet of a ReLL Description for the School of Information Web Site}\label{ReLL_iSchool}
\end{center}
\end{figure}

\begin{figure*}[tb]
\begin{center}
\includegraphics{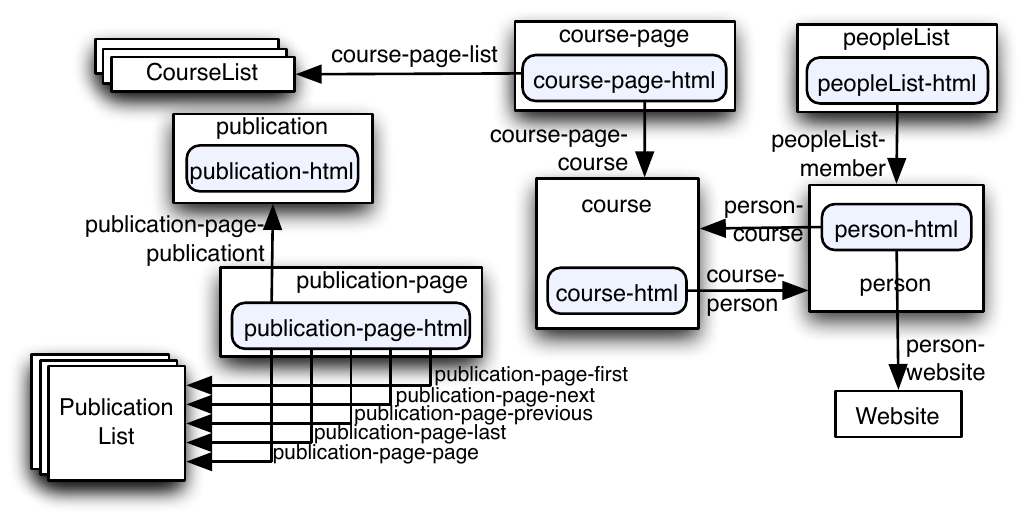}
\caption{A ReLL based model for the School of Information Web site}\label{school-model}
\end{center}
\end{figure*}

ReLL descriptions are XML documents created according to the ReLL schema. Figure~\ref{school-model} shows part of the description of the service that publishes interlinked HTML pages on the Web site of the School of Information at UC Berkeley. This service is a Web \emph{Content Management System (CMS)} publishing interlinked Web pages. Even though we have no control over that CMS, we can describe the set of paged made available my it as a set of interlinked resources. Figure~\ref{ReLL_iSchool} shows only one resource, a person's page that links to the courses taught by that person and the person's personal home page, and in general, we did not strive for completeness in that demonstration scenario.

Rectangles in Figure~\ref{school-model}, which is a graphic representation of the ReLL model, represent resources, and rounded rectangles their \emph{representations}. Arrows between representations and resources are links. Collections are sets of resources of the same type (e.g., \emph{courseList}). The collection itself is a conceptual resource with no representation or URI. The rectangle labeled as \emph{``Website"} corresponds to a resource that is out of the scope of the service, it thus has no representation. The link that leads to this resource from a \emph{``person-html"} representation does not include the \emph{target} as show in Figure~\ref{ReLL_iSchool}.

\section{Harvesting RDF from Resources}\label{harvesting}

\begin{figure}[tb]
\begin{center}
\includegraphics[width=0.85\textwidth]{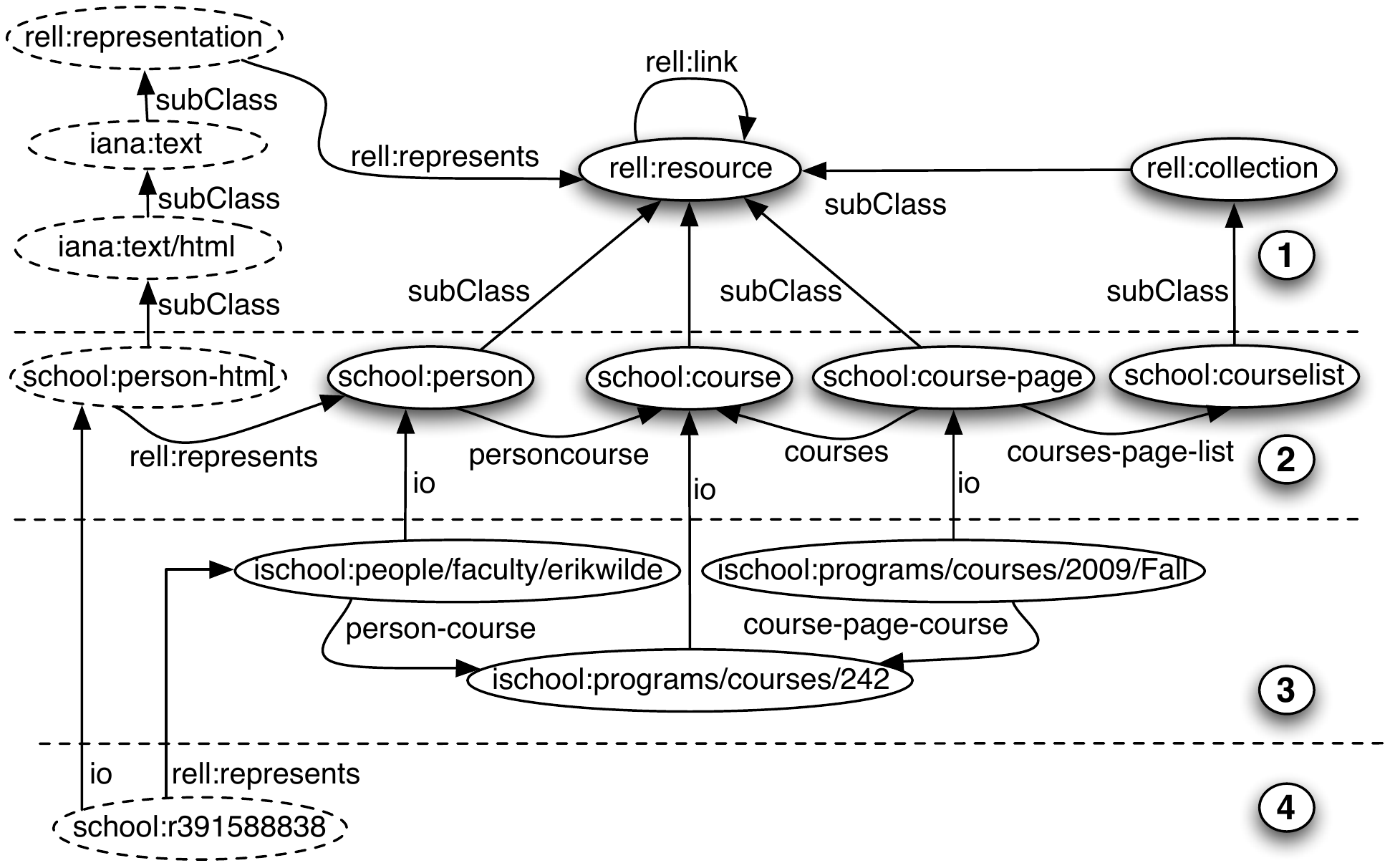}
\caption{Semantic model for the REST Services Metamodel}\label{rell-ontology}
\end{center}
\end{figure}

Figure~\ref{rell-ontology} shows the ReLL metamodel mapped to a RDF/OWL semantic model. We consider four layers in the semantic model. Layer 1 corresponds to the \emph{upper ontology} that describes general REST semantics, layer 2 corresponds to a domain ontology describing a specific RESTful service (e.g., the Web site of the School of Information). Layer 3 contains the data discovered when crawling the service, that is, the REST resources. Layer 4 contains information about concrete representations that have been used for establishing the resource relationships of Layer 3.

\begin{figure}[tb]
\begin{center}
\includegraphics[width=\textwidth]{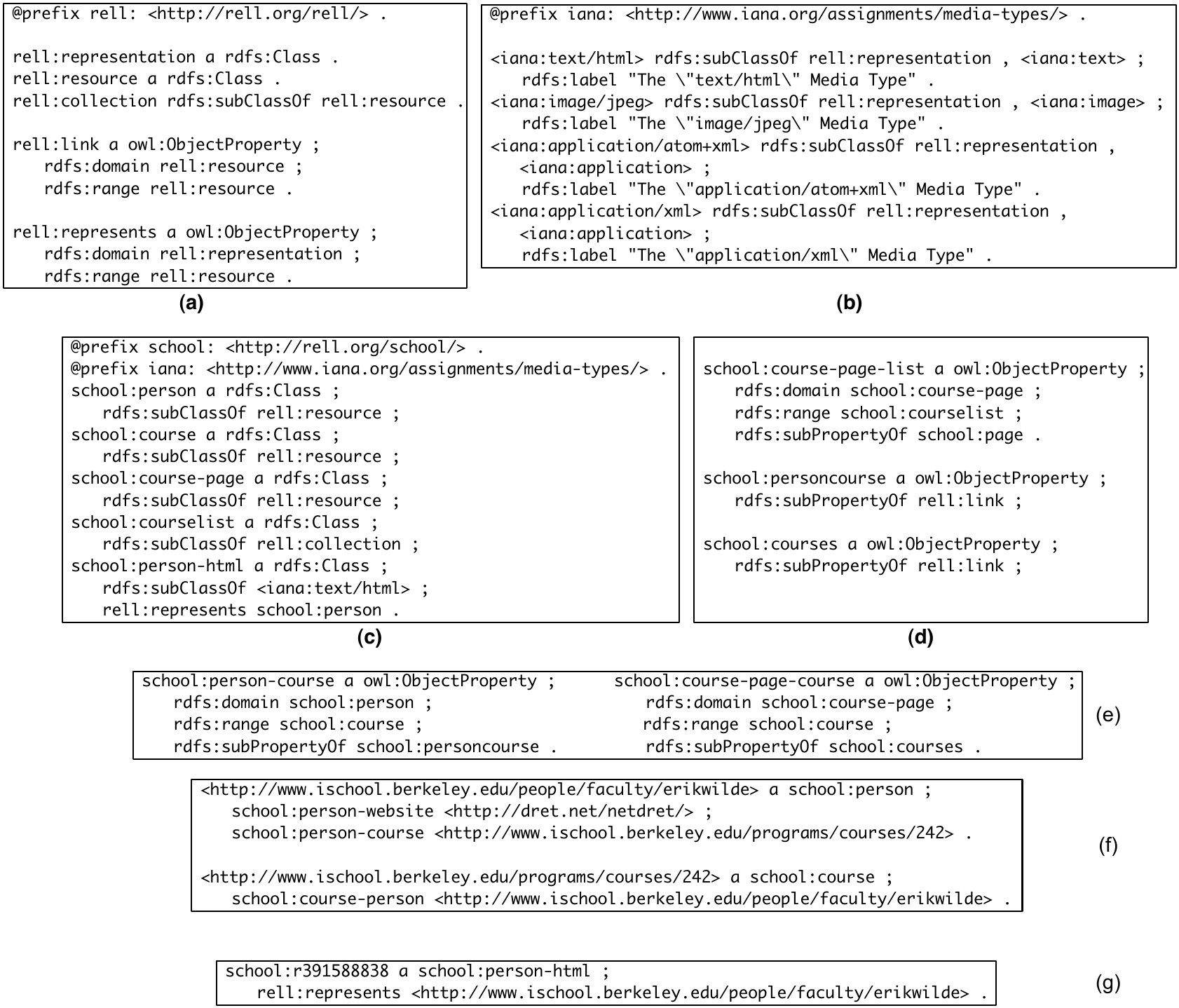}
\caption{N3 notation snippets of RDF triples generated for REST services}\label{triples}
\end{center}
\end{figure}

The elements with dashed outlines in Figure~\ref{rell-ontology} correspond to the following strategy: The \emph{resource}, \emph{representation} and \emph{collection} elements in the ReLL metamodel correspond to classes of the upper ontology in layer 1. \emph{Selector} is not considered since it is a medium for selecting elements embedded in the Web representation content. We have made a difference between how links \emph{work} in the realm of the Web, and what links \emph{assert} in the realm of the Semantic Web. In the former case, links are contained in representations and thus relate representations to resources; in the latter case, representations are considered as provenance, merely indicating the source of the crawled information. Hence, in our model, representations \emph{represent} resources, and \emph{links} relate resources to resources. Figure \ref{triples}a presents the upper ontology in N3 format.

Layer 2 corresponds to the domain ontology describing a REST service. REST resources do not have types, only their representations have media types. In the Semantic web, resource types allow a more expressive model that facilitates tasks such as querying and inferencing. ReLL \emph{resources} and \emph{representations} are translated into subclasses of the upper ontology\footnote{For representations, the upper ontology contains all standardized media types from the IANA registry as classes.} and \emph{link types} become subproperties of the \code{rell:link} property. Figure~\ref{triples} presents a subset of the generated classes (c) and properties (d). Labels and comments are omitted and the subset is limited to the example shown in Figure~\ref{rell-ontology}. Collections are classes (e.g., courselist) and their members are explicitly linked by means of the \emph{link type}.

\begin{figure}[tb]
\begin{center}
\includegraphics[width=\textwidth]{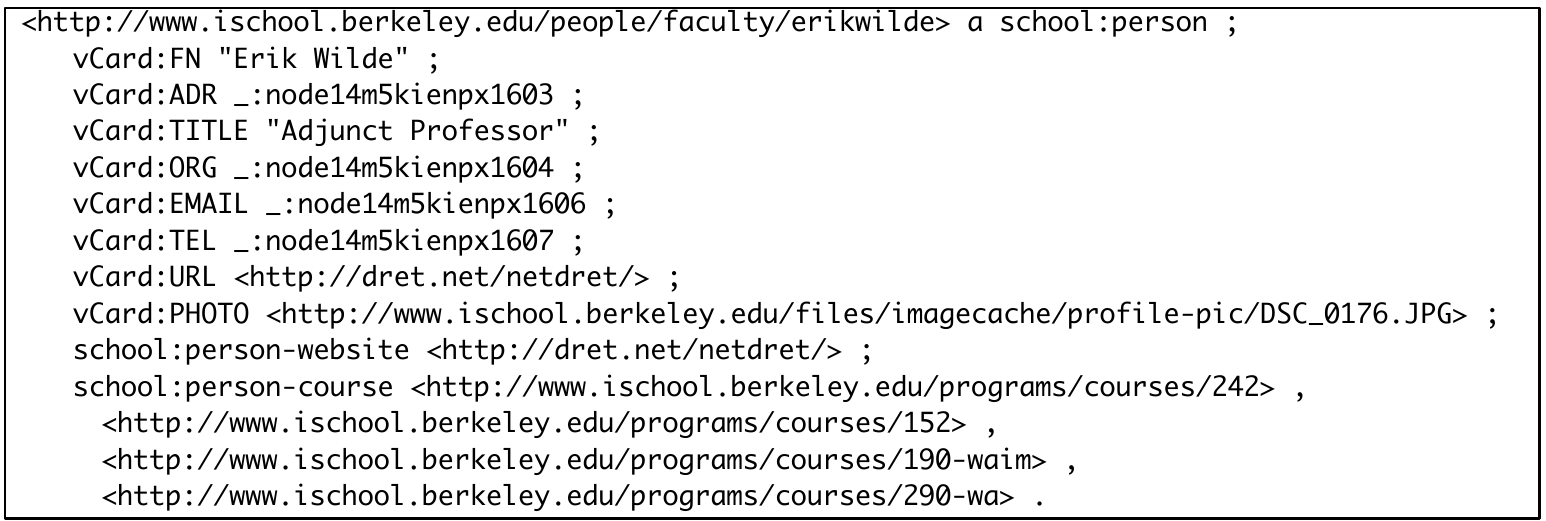}
\caption{An individual's properties in N3 notation}\label{list-attribute}
\end{center}
\end{figure}

Layer 3 contains REST resources modeled as individuals of the domain classes (\code{rdf:type}). Since REST resources have unique identifiers, we maintain them for identifying the individuals. We create subproperties of the domain relationships for relating actual individuals and use the \emph{link identifiers} for naming them. Figure~\ref{triples}e describes the case for \code{person-course} and \code{course-page-course} properties. We also restrict their domain and range. Figure~\ref{triples}f presents the triples asserted for the individuals shown in Figure~\ref{rell-ontology} as well as their relationships. The REST resources themselves are transformed to RDF following a GRDDL approach. Figure~\ref{list-attribute} shows the attributes obtained for individuals of type \code{person}. Notice that it is possible to annotate the relationships between the REST resource (\code{erikwilde}) and its attributes. In the figure these relationships are annotated with \emph{vCard}, but other information models can be used.

Layer 4 represents provenance. Triples obtained in layer 3 are produced when crawling a REST service, that is, a REST resource URI was dereferenced and its representation obtained, and the links to related resources were retrieved from the representation by evaluating the XPath expressions indicated by ReLL selectors. This process generates a number of triples that are tied together as a named graph~\cite{car05}, where the name is a generated ID (e.g., \code{school:r391588838}). The ID identifies a representation which is also asserted as an instance of a media type (and hence an instance of a \code{rell:representation}). A relationship between the representation and the resource (\code{rell:represents}) is also asserted (Figure~\ref{triples}g). We have modeled concrete representations as classes that derive from abstract representations, which correspond to \emph{media types}. Figure~\ref{triples}b presents an example of the media type classes describing \code{text/html}, \code{image/jpeg}, \code{application/atom+xml}, and \code{application/xml} media types.

Media types are annotated in the ReLL descriptions (Figure~\ref{ReLL_iSchool}). Abstract representations are supported as classes that serve as the basis for other abstract or concrete representations. In Figure~\ref{rell-ontology}, the \code{text} media type is an abstract representation that serves as the basis for the \code{text/html} media type, which is also an abstract representation and serves as the basis for a concrete representation, that is an HTML page describing a person.

\section{Composition as a Service}\label{composition}

\begin{figure}[tb]
\begin{center}
\includegraphics[width=\textwidth]{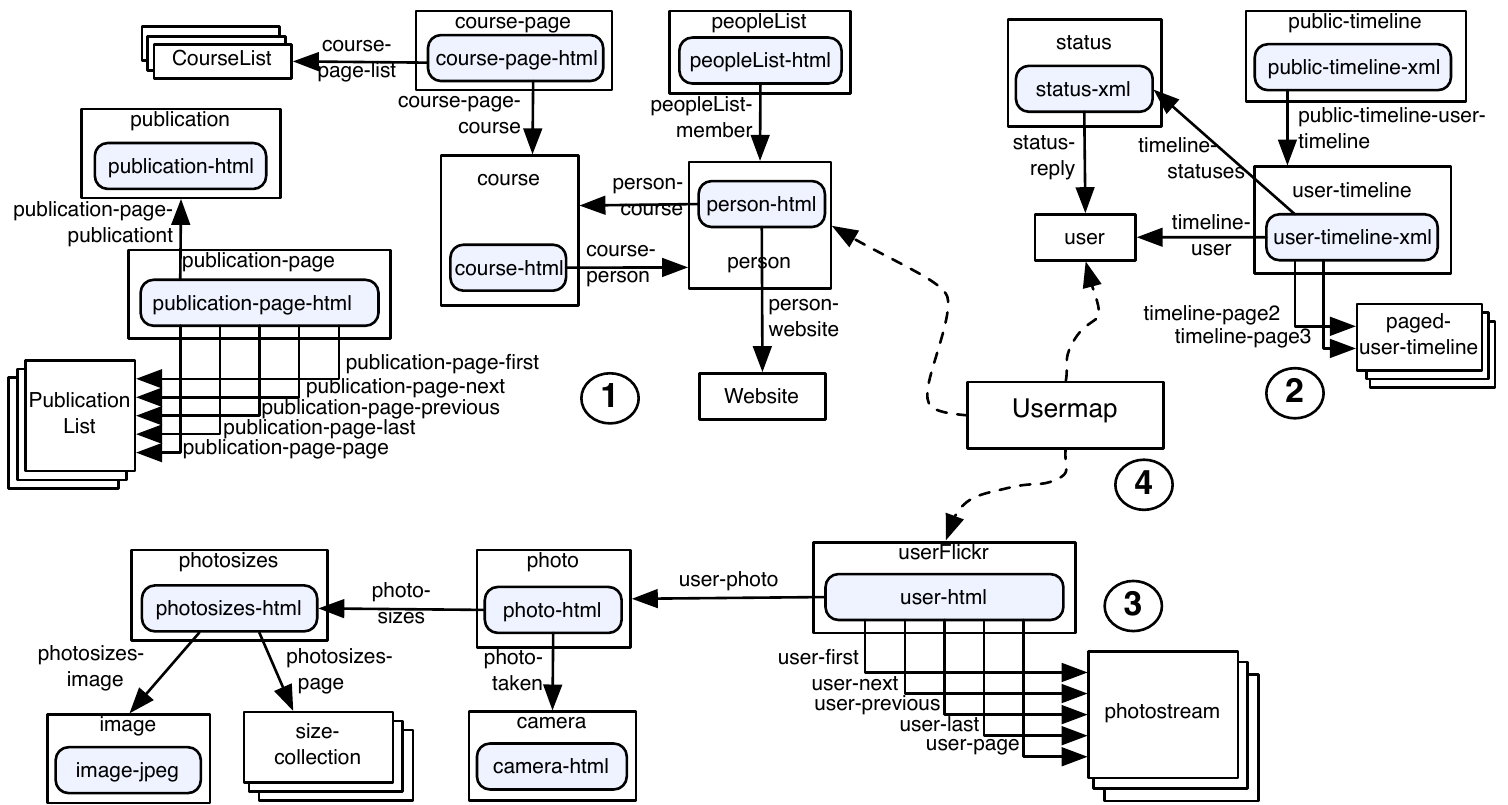}
\caption{Composing four REST services, School, Twitter, Flickr and UserMap}\label{composite}
\end{center}
\end{figure}

Three other REST services besides the School of Information Web site have been described using ReLL: a service that publishes Atom feeds through a REST API corresponding to Twitter, a service that publishes interlinked HTML pages corresponding to the Web site of Flickr, and a service that publishes an XML document listing the identities of the users of these applications (this service provides the ``glue'' for associating different user identities in the other services). Figure~\ref{composite} presents ReLL models for each of these applications labeled as 1 (School), 2 (Twitter), 3 (Flickr) and 4 (UserMap) respectively.

\begin{figure}[tb]
\begin{center}
\includegraphics[width=\textwidth]{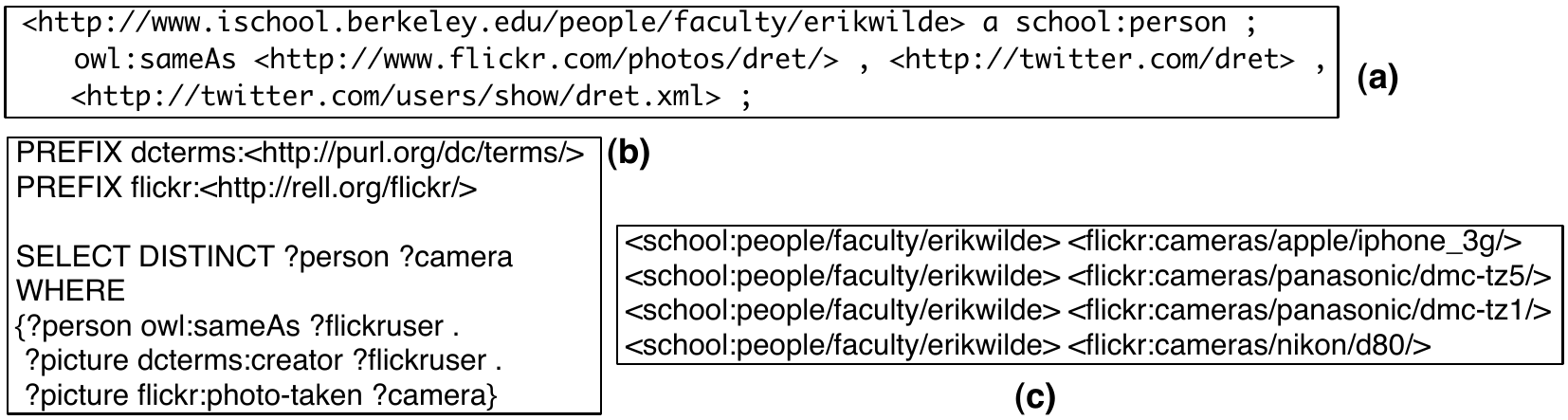}
\caption{N3 notation snippets for the composition scenario}\label{sameas}
\end{center}
\end{figure}

The UserMap service publishes an XML document that groups together user identities for all three services (School, Twitter, and Flickr), and thus the composition accomplished by the UserMap service is simply another service, one that provides the ``glue'' that is required to connect the aggregated services into a connected composition of resources. For the RDF mapping, an XSLT transforms the XML document into triples that establish the equivalences between these resources through \code{owl:sameAs} assertions. Figure~\ref{sameas}a presents the triples for one user. Notice that for the Twitter case, two URIs are considered, the URI of the resource as obtained from the REST API, and the URI of the resource as obtained from the Twitter Web site. The proposed approach allowed us to perform SPARQL queries that cover many REST services, as shown in Figure~\ref{sameas}b. The query retrieves the list of distinct \code{flickr:cameras} for all Flickr photos of a \code{person}. Figure~\ref{sameas}c presents a snippet of the results for the user \code{erikwilde}.

Figure~\ref{sameas}c can serve as a demonstration of the overall value of our approach. The fact that a particular URI identifies a person is established by using the information available on the School of Information Web site. The fact that this person is the \code{owl:sameAs} some Flickr user has been established by the UserMap service, where a table of associated user identities has been transformed into RDF by a GRDDL XSLT transform. That Flickr user's photos have been crawled by using the ReLL description of the Flickr service and transforming this information into RDF. Finally, the metadata about the particular camera used for each photo has been extracted from each photo's information page, again through GRDDL XSLT. This means that the only information required to get to a connected graph that can be queried as shown in Figure~\ref{sameas}b are ReLL descriptions, XSLT transforms for some resource types, and a composition service.

\section{Implementation and Results}\label{implementation}

\begin{figure}[tb]
\begin{center}
\includegraphics{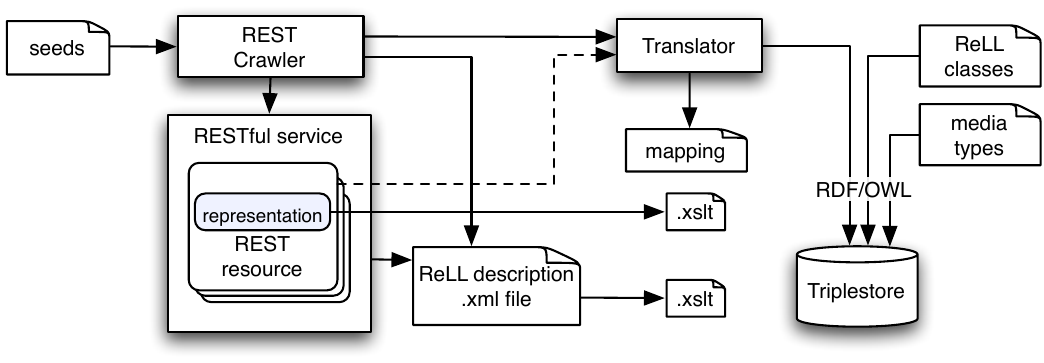}
\caption{Implementation}\label{fig-implementation}
\end{center}
\end{figure}

Figure~\ref{fig-implementation} shows the components for harvesting triples from RESTful services. Services are described by ReLL generating XML documents that direct the actions of a Web crawler. Since REST services do not have ``endpoints'', a list of seed URIs is required. There is no guarantee that the whole resource graph will be covered since this depends on how well the resources of a service are connected. While crawling, a translator component is invoked for generating RDF triples. Translation is optional, a property file gives the translator the mappings between the resource's type (e.g., \code{person}) that is required to translate to RDF, and XSLT code. XSLT transforms are defined for ReLL description in order to generate domain triples (Layer 2), and for the representation contents in order to generate attributes (Layer 3). Resource URIs, resource types and link types are passed to the translator in order to assert individuals and properties. We use Sesame 2.0 as triple store and the system is implemented in Java.

\begin{figure}[tb]
\begin{center}
\includegraphics[width=\columnwidth]{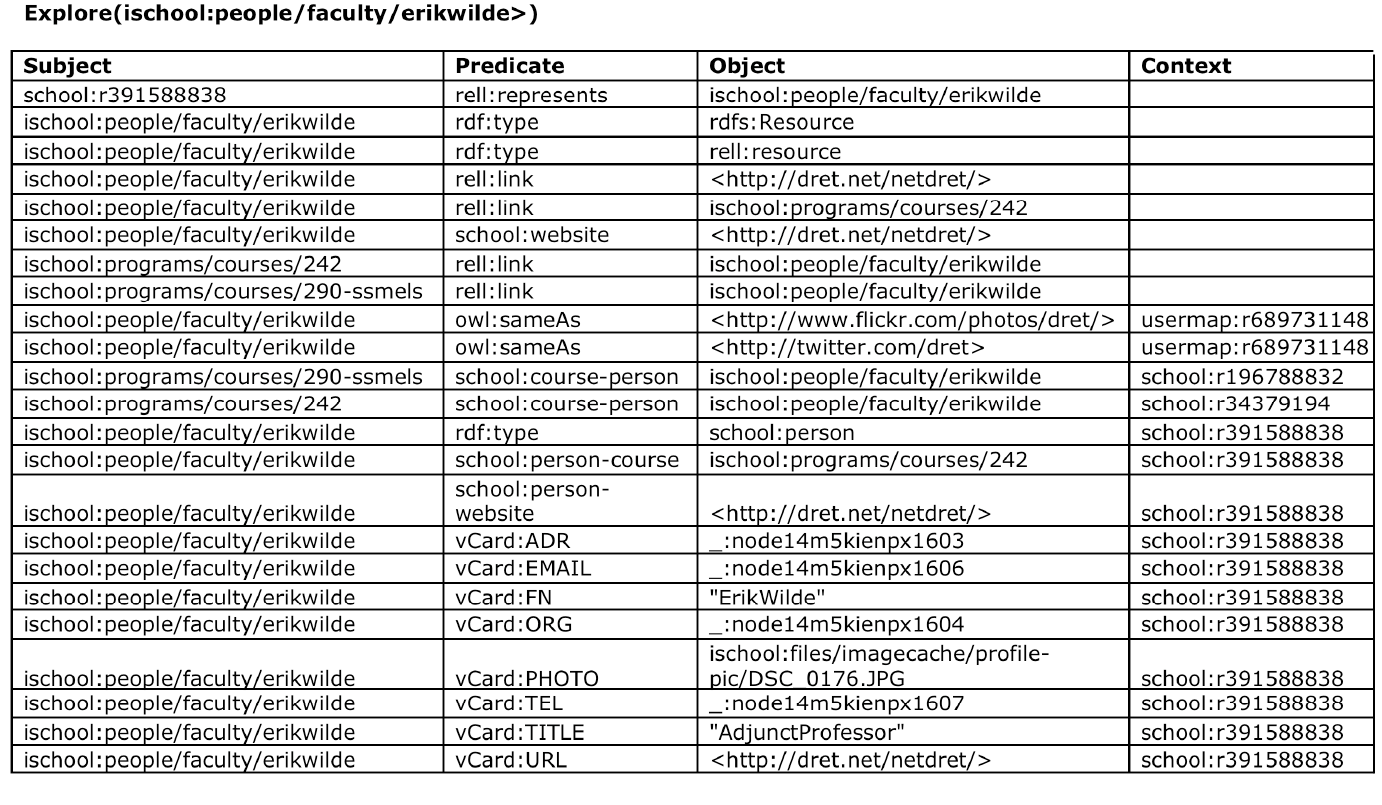}
\caption{Description for one User in the Composite Service}\label{results1}
\end{center}
\end{figure}

Sesame supports named graphs as context, which is a fourth component that can be added to a triple, and it is possible to treat that fourth component as a semantic resource (\emph{rdf:Resource}). There is no need to modify the triples, since context is manipulated trough Sesame's Java interface. Triples corresponding to the ``upper" ontology and media types taxonomy are asserted into the triple store directly. Figure~\ref{results1} presents the results of querying the triple store for \code{ischool:people/faculty/erikwilde} through the Sesame export functionality. In the first row the representation element is shown and is presented as the context element (quad) in the following items. For the triples in Layer 3 (such as those with \code{school:course-person}) a new triple is inferred, since the property is a subproperty of \code{rell:link}, and in this case there are not a context elements.

\section{Conclusions}\label{conclusions}

In this paper we propose a method and implementation for harvesting triples from services and service compositions that follow the REST architectural principles. Most Web sites fall into that category, which implies that a large dataset may be available to be translated to RDF. We propose a lightweight approach that places a strong emphasis on flexibility by decoupling the main components. That is, REST services do not require modifications and do not depend on existing ReLL descriptions for our approach, and ReLL descriptions do not contain information for the translation to RDF, they can be used independently for other purposes such as documentation or as the starting point for a service contract. Mapping is isolated in one layer and as far as possible can be configured by dynamic files. We believe that this approach is sufficiently generalized to be applied to various data sources provided that they follow the REST architectural principle.

Currently, we are not providing information about the representations in the RDF data, but we intend to continue this work in that direction. We consider that they may include time stamps indicating the last time when the resource was crawled, entity tags (ETags) served by Web servers indicating whether the resource has changed since the last retrieval, or other HTTP metadata.

The two most challenging research questions we are facing is whether we can extend the architecture to support incremental harvesting, so that large services do not need to be completely recrawled for getting updates, and whether we can further extend this direction by supporting ``on-demand service access'', so that queries into the triple store are actually mapped to live services instead of using harvested triples. In that latter case, the triple store would essentially become a cache instead of a separate dataset, and the problem of deciding whether to serve harvested RDF or whether to selectively recrawl the underlying services might nicely translate into the more general problem of how to efficiently describe and use cacheable services on the Web.

\bibliographystyle{dret}
\bibliography{dret,ralarcon}

\end{document}